\documentclass[letterpaper, 10 pt, journal, twoside]{IEEEtran}

\usepackage{graphicx}
\usepackage{cite}
\usepackage{color}
\usepackage{xcolor}
\usepackage{soul}
\usepackage{amssymb}  
\usepackage{amsmath, bm, amsfonts}
\usepackage{subfigure}
\usepackage{threeparttable}
\usepackage{tabularx}
\usepackage{multirow}
\usepackage{booktabs}
\usepackage[linesnumbered,ruled,vlined]{algorithm2e}
\usepackage{algorithmicx}
\usepackage{enumerate}
\usepackage{ulem}
\usepackage{url}
\newcommand{\magenta}[1]{\textcolor{magenta}{#1}}

\begin{document}
\title{Learning Efficient Policies for Picking \\Entangled Wire Harnesses: An Approach to\\Industrial Bin Picking}

\author{Xinyi Zhang$^{1}$, Yukiyasu Domae$^{2}$, Weiwei Wan$^{1}$, and Kensuke Harada$^{1,2}$
\thanks{Manuscript received: July 10, 2022; Revised October 6, 2022; Accepted October 29, 2022.}
\thanks{This paper was recommended for publication by Editor Hong Liu upon evaluation of the Associate Editor and Reviewers' comments.
This work was supported by Toyota Motor Corporation.
} 
\thanks{$^{1}$Xinyi Zhang and Weiwei Wan are with the Graduate School of Engineering Science, Osaka University, Toyonaka, Osaka 560-0043, Japan (e-mail: xinyiz0931@gmail.com; wanweiwei07@gmail.com).}
\thanks{$^{2}$Yukiyasu Domae is with the National Institute of Advanced Industrial Science and Technology, Tokyo, Japan (e-mail: domae.yukiyasu@aist.go.jp).}
\thanks{$^{1,2}$Kensuke Harada is with the Graduate School of Engineering Science, Osaka University, Toyonaka, Osaka 560-0043, Japan, and also with the National Institute of Advanced Industrial Science and Technology, Tokyo, Japan (e-mail: harada@sys.es.osaka-u.ac.jp).}
\thanks{Supplementary material, code and video can be found at \\\protect\url{https://github.com/xinyiz0931/aspnet}. }
\thanks{Digital Object Identifier (DOI): see top of this page.}
}


\markboth{IEEE Robotics and Automation Letters. Preprint Version. Accepted November, 2022}
{Zhang \MakeLowercase{\textit{et al.}}: Learning Efficient Policies for Picking Entangled Wire Harnesses} 

\maketitle

\begin{abstract}
    Wire harnesses are essential connecting components in manufacturing industry but are challenging to be automated in industrial tasks such as bin picking. They are long, flexible and tend to get entangled when randomly placed in a bin. This makes it difficult for the robot to grasp a single one in dense clutter. Besides, training or collecting data in simulation is challenging due to the difficulties in modeling the combination of deformable and rigid components for wire harnesses. In this work, instead of directly lifting wire harnesses, we propose to grasp and extract the target following a circle-like trajectory until it is untangled. We learn a policy from real-world data that can infer grasps and separation actions from visual observation. Our policy enables the robot to efficiently pick and separate entangled wire harnesses by maximizing success rates and reducing execution time. To evaluate our policy, we present a set of real-world experiments on picking wire harnesses. Our policy achieves an overall 84.6\% success rate compared with 49.2\% in baseline. We also evaluate the effectiveness of our policy under different clutter scenarios using unseen types of wire harnesses. Results suggest that our approach is feasible for handling wire harnesses in industrial bin picking. Supplementary material, code, and videos can be found at \magenta{\url{https://xinyiz0931.github.io/aspnet}}. 
\end{abstract}




\section{Introduction}\label{sec:intro}
    
\begin{figure}[t] 
    \centering
    \includegraphics[width=\linewidth]{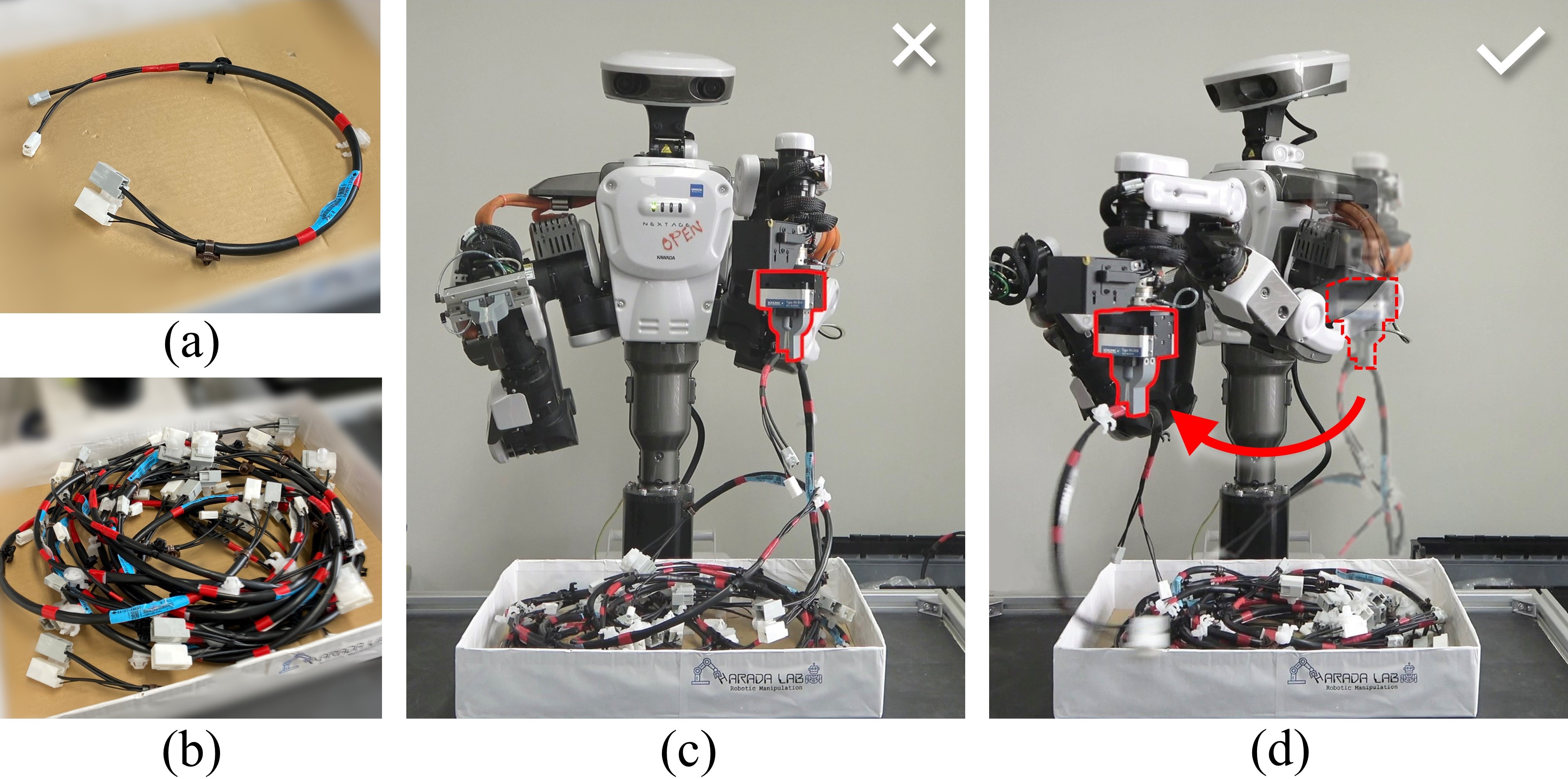}    
    \caption{(a-b) Wire harnesses are composed of both deformable and rigid components. They get entangled easily in clutter and their length may exceed the robot arm's reach areas. (c) Directly lifting a wire harness causes entanglement. (d) We learn a bin picking policy to efficiently extract an entangled wire harness from an unstructured bin. }
    \label{fig:teaser} 
\end{figure}

    \IEEEPARstart{B}{in} picking is a vital task in manufacturing industry that enables a robot to pick objects randomly placed in a bin. If we try to automate an assembly process without using bin picking, we need to prepare a large amount of parts feeders according to the number of assembly parts. Although robotic bin picking has been researched for decades \cite{domae2014fast,liu2012fast,kirkegaard2006bin,harada2016initial,matsumura2018learning,tachikake2020learning}, some objects (e.g., wire harnesses) can still be challenging when automating this process. A wire harness is an indispensable component used in almost every electric drive product. Fig. \ref{fig:teaser}(a) shows its appearance. It comprises a group of bundled wires and multi-conducted connectors and is used for transmitting signals and power. The structure of a wire harness also poses challenges in robotic bin picking: (1) The existence of both deformable and rigid components makes them easily form an entangled clutter in the bin; (2) The complex geometries and deformable nature cause difficulties in 3D modeling; (3) The length of a wire harness often exceeds the operation range of a robot, making it difficult to extract one from the bin. To successfully perform bin picking using wire harnesses, the robot must be equipped with the capability of effectively isolating each from the entanglement. For this reason, the manufacturing industry still relies on human workers to grasp and separate entangled wire harnesses. Therefore, developing an intelligent system to automate this process is highly demanded. 
    
    Existing works on industrial bin picking have primarily focused on rigid parts. These methods grasp objects by avoiding collisions in highly cluttered environments \cite{domae2014fast,harada2016initial,matsumura2018learning,tachikake2020learning,dupuis2008two,buchholz2014combining}. For picking simple shaped objects, the robot usually lifts the target in the vertical direction after a successful grasp. Different from those objects, wire harnesses involve complex entanglement when randomly placed in a bin. Besides, they are much longer than the rigid parts already automated in bin picking. The physical reach range of the robot in a bin picking working cell is limited for completely lifting them. Simply adapting the existing bin picking strategies shows unsatisfied performance (see Fig. \ref{fig:teaser}(c)). Previously, some studies have addressed the entanglement problems but for picking curved rigid parts by avoiding the potentially tangled parts \cite{matsumura2019learning,zhang2021topological}. However, there remain problems for densely cluttered wire harnesses where the bin often contains no isolated objects as Fig. \ref{fig:teaser}(b) shows. Motivated by modeling and manipulating deformable linear objects, studies on visually processing wire harnesses start with segmenting or generating synthetic data for wire harnesses with pure linear shapes \cite{guo2022visual}, \cite{caporali2022fastdlo}. For those with complex geometries, obtaining precise models or training in simulation remains difficult. Alternatively, employing a real robot to collect large-scale data is time-consuming. Annotating ground truth labels is also challenging due to the lack of entanglement metrics.

    In this paper, we (1) design an effective motion to untangle wire harnesses in clutter and (2) learn a policy to perform bin picking tasks with higher success rates and lower execution time. The key components of our system are: 
    
\begin{itemize}
    \item We propose a post-grasping action to untangle wire harnesses. Instead of lifting in the vertical direction, the robot separates the entangled objects in the horizontal direction. The action continuously follows a circle-like trajectory to extract the target within the limited robot's reach range. Fig. \ref{fig:teaser}(d) shows this process. 
    \item We learn a bin picking policy to infer an optimal grasp and a post-grasping action from a depth image. Our policy can prioritize grasping the untangled objects, avoid grasping at the bad positions (e.g., the ends of the object) and reason the extracting distance to reduce the execution time for a successful picking. Additionally, we train the policy with real-world data by leveraging active learning for satisfying convergence. 
\end{itemize}
    
    Our contributions are three-fold. (1) We develop a unique bin picking system that can disentangle wire harnesses from dense clutter. (2) Instead of lifting the target in the vertical direction after grasping, our policy proposes to simultaneously lift and move in the horizontal direction for separating wire harnesses. (3) We learn a policy using real-world data to infer the optimal actions, which further improves bin picking efficiency. Real-world experiments suggest our policy can significantly improve the average success rates and reduce operation time compared with baselines. 


\section{Related Work}\label{sec:rw}
    
\subsection{Industrial Bin Picking}\label{subsec:rw-bin}
    
    Industrial bin picking has been developed for decades. Prior works have primarily focused on model-based approaches such as 3D object localization or 6D pose identification \cite{liu2012fast,choi2012voting,yang2021probabilistic} and grasp planning \cite{dupuis2008two,buchholz2014combining,harada2013probabilistic}. Alternatively, model-free methods do not require known object information and can produce grasp poses for novel objects. Domae et al. \cite{domae2014fast} proposed to plan grasps considering collisions between the gripper and the objects from a single depth image. Several works leverage deep learning to mitigate the challenges of complex physical interactions and environment uncertainties. Mahler et al. \cite{mahler2017learning} trained a model from synthetic data to produce collision-free grasps for daily objects. Matsumura et al. \cite{matsumura2018learning} proposed a learning-based method to plan robust grasps for industrial parts. However, there remain challenges in handling difficult objects. Recently, several works tackled the challenges of those objects which are (1) difficult to be recognized, e.g., transparent or shiny objects \cite{tachikake2020learning,li2022sim}, (2) difficult to perform grasping, e.g., thin and elliptical objects \cite{tong2021dig,morino2020sheet}, and (3) involved with rich physical interactions, e.g., tangled-prone objects \cite{matsumura2019learning,zhang2021topological}. So far, these approaches have focused on rigid objects. Leao et al. \cite{leao2020detecting} proposed a method to pick up soft tubes by fitting shape primitives to clutter, but it does not work on dense clutter or objects with irregular shapes. Objects with non-rigid nature and complex geometries like wire harnesses are relatively unexplored and remain challenging in the industrial bin picking domain. 

\subsection{Deformable Object Manipulation}\label{subsec:rw-deform}

    Deformable object manipulation has primarily focused on two object classes: 1D (cable, rope) and 2D (fabric, cloth). Several studies adopt specially designed motion primitives to accomplish various manipulation tasks such as knot tying/untying \cite{yamakawa2008knotting,lui2013tangled, grannen2021untangling}, spreading cloth \cite{ha2022flingbot} or whipping ropes \cite{chi2022irp}. Using deformable and long objects in industrial bin picking poses new challenges. The cluttered scenes are more complex due to the entanglement issues caused by their deformable nature. Ray et al. \cite{ray2020robotic} proposed to untangle herbs from a pile using a two-finger gripper. Takahashi et al. \cite{takahashi2021target} proposed a learning-based separation strategy for grasping a specified mass of small food pieces. Although some works have addressed the factory automation problems for wire harnesses \cite{guo2022visual,jiang2011robotized,zhou2020practical}, robotic wire harnesses picking is less studied. In this paper, we propose a novel and efficient bin picking strategy to deal with wire harnesses. 
    

\section{Motion Primitives for Disentangling}\label{sec:pull}

    When a robot tries to isolate small and rigid objects from a bin, it can lift them in a vertical direction after a successful grasp. However, this movement is insufficient for isolating long and flexible objects like a wire harness, whose length exceeds the bin picking workspace. To extract such objects, the required motion primitives must be designed to (1) provide enough space for effectively disentangling long objects and (2) handle various tangle patterns. Instead of directly lifting, the horizontal movement of the gripper can help pull the target object out. The possible positions of the gripper should also remain in the outer part of the parts bin during disentangling. In the end, we propose two motion primitives for effectively disentangling a long and flexible object: 
    
\begin{figure}[h] 
    \centering
    \includegraphics[width=\linewidth]{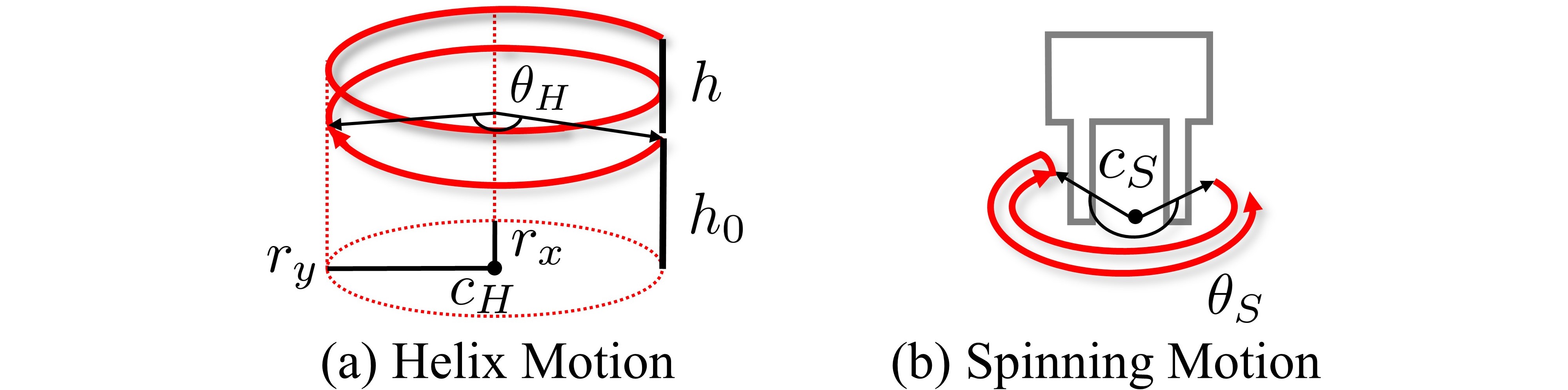} 
    \caption{(a)  Helix motion primitive $\psi_H=(H,\theta_H)$ where the helix trajectory is defined as $H=(c_H,r_x,r_y,h_0,h)$. (b) Spinning motion primitive $\psi_S=(\theta_S,c_S)$.}
    \label{fig:action-traj} 
\end{figure}

\begin{figure}[t] 
    \vspace{0.1cm}
    \centering
    \includegraphics[width=\linewidth]{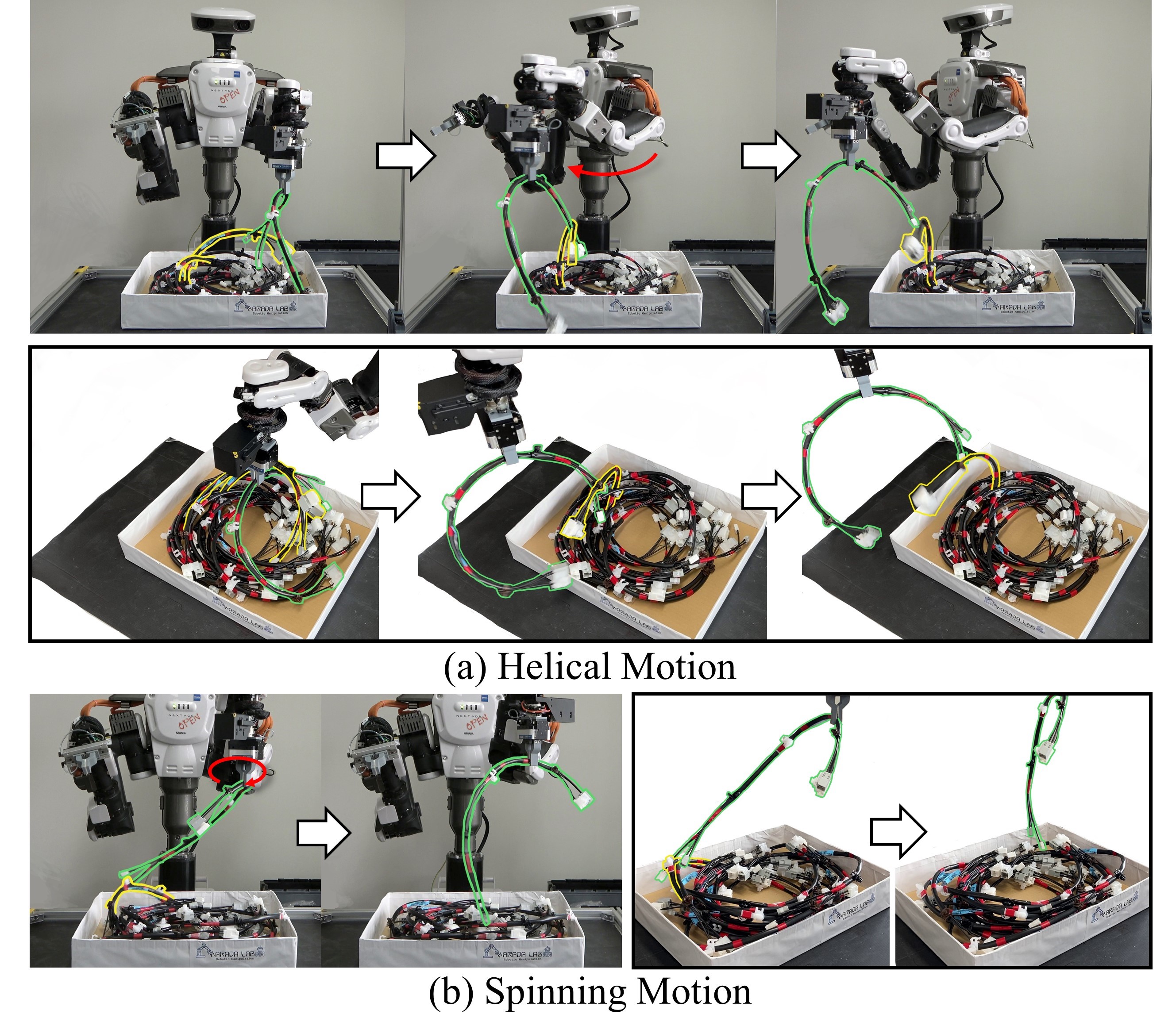} 
    \caption{The proposed motion primitives can handle two properties of wire harnesses: tangle-prone and length. (a) The robot separates an entangled wire harness from a gentle angle following a helix trajectory. (b) A spinning motion is performed when the target's connectors slightly hang on the other objects. }
    \label{fig:action-pull} 
\end{figure}

    \noindent
    \textbf{Helix motion:} $\psi_H=(H, \theta_H)$ where $H$ denotes the helix trajectory represented by $(c_H , r_x, r_y, h_0, h)$ and $\theta_H$ denotes the execution angle following the trajectory (see Fig. \ref{fig:action-traj}(a)). $c_H$ denotes the base center of $H$ and $r_x,r_y$ constrain the  smallest and largest radius from the center. $h$ denotes the height of $H$. The helix starts after the gripper lifts the target and reaches $h_0$. The stop point of the helix is determined by the execution angle $\theta_H$. It is a post-grasping motion where the gripper simultaneously lifts and pulls following a helix-like trajectory. Let the gripper move around the bin while holding an entangled object. Part of this object is also moving outside the bin. When the gripper continuously moves like drawing circles, the grasped object can be disentangled softly along a side angle. Fig. \ref{fig:action-pull}(a) shows that this movement provides adequate space to pull the target (green) out of the entangled objects (yellow). Meanwhile, we also observe that other entangled objects remain in the bin during or after this process, making the workspace clean for the next picking.\\
    \textbf{Spinning motion:} $\psi_S=(c_S,\theta_S)$ where $c_S$ denotes the position of the gripper tip and $\theta_S$ denotes the one-way rotation angle of the spinning (see Fig. \ref{fig:action-traj}(b)). The robot performs a two-way spinning about the axis that is vertical to the robot workspace. The gripper spins to handle the entanglement that may be occluded from the observation. As Fig. \ref{fig:action-pull}(b) shows, when the rigid components of the wire harness still slightly hang on the others after the helix motion, an extra spinning can help separate them with less execution time. It can also handle the length of a wire harness by extracting it inside a limited working cell.

\section{Learning Bin Picking Policies}\label{sec:method}
\begin{figure}[t] 
    \vspace{0.1cm}
    \centering
    \includegraphics[width=\linewidth]{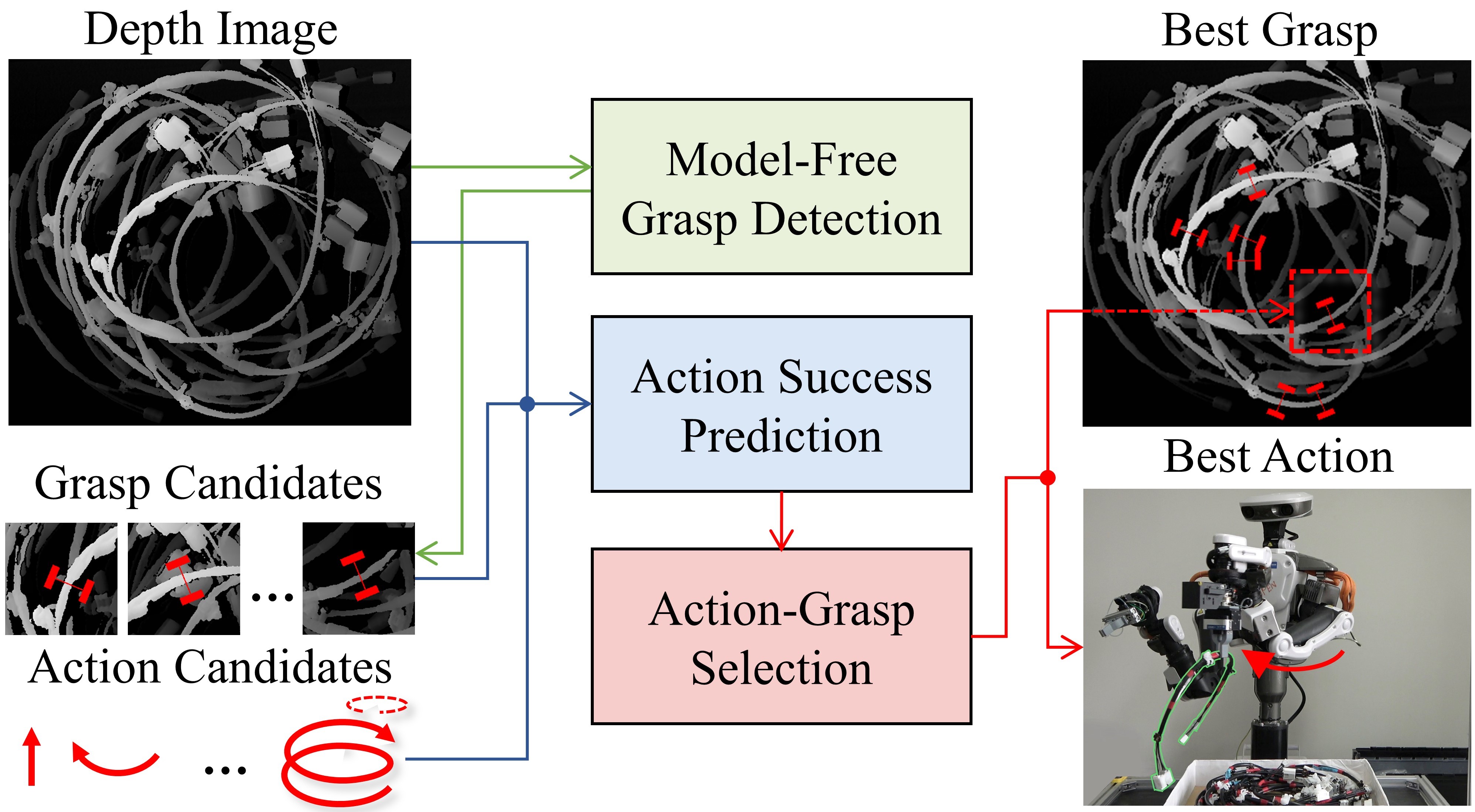} 
    \caption{Overview of our policy. Given a depth image of an unstructured bin, Model-Free Grasp Detection module samples a set of non-collision grasp candidates. Then, Action Success Prediction module takes a depth image, grasp candidates and action candidates as input and evaluates the success possibility for each action-grasp pair. Finally, Action-Grasp Inference module ranks these pairs and outputs the optimal action and grasp. }
    \label{fig:pipeline} 
\end{figure}

    The goal of our bin picking policy is to pick up a single wire harness at a time by inferring the optimal grasp and action from current entanglement situation. If the scene contains isolated objects, the robot prefers directly lifting them after grasping. Otherwise, the robot can infer disentangling actions and grasp poses to extract the target from the bin. Given a top-down depth image $o$ as observation, we formulate our bin picking policy $\pi$ with a trained model parameterized by $\tau$ using: 
\begin{equation}
    a^*,g^*=\pi_{\tau}(o)
\end{equation}
    where the outputs are an action $a^*$ and a grasp $g^*$ with the maximal task effectiveness. The action $a$ comprises the proposed motion primitives. Fig. \ref{fig:pipeline} shows the three essential modules in our policy: \\
    \textbf{Module I. Model-Free Grasp Detection:} A grasp detection algorithm using a depth image without object models.\\
    \textbf{Module II. Action Success Prediction (ASP):} A trained model using real-world data that predicts the success possibilities of the disentangling actions. \\
    \textbf{Module III. Action-Grasp Inference:} A method to infer the action-grasp pair with the highest effectiveness using the trained ASP model.

\subsection{Model-Free Grasp Detection}\label{subsec:method-grasp}
    We select Fast Graspability Evaluation (FGE) \cite{domae2014fast} - a model-free approach to detect collision-free grasps. FGE calculates pixel-wise graspability scores by convoluting a gripper's template of contact and collision areas with the input depth map. A grasp composes a pixel location $g=(u,v)$ on the depth map and a rotation angle $\phi$ indicating the gripper's orientation. We transform $(u,v,\phi)$ to the grasp with four degrees of freedom $(g_x,g_y,g_z,g_\phi)$ denoting the grasp point and the gripper's orientation at the robot coordinate frame. This module outputs a set of grasps ordered by their FGE scores. 
    
\subsection{Action Success Prediction (ASP)}\label{subsec:method-action}
    
    \textit{1) Action Formulation:}  We formulate each disentangling action $a$ with a motion scheme $\psi$ and two parameters as follows: 
\begin{equation}
    a = (\psi,\theta_H,\theta_S) \; | \; \psi=\{\psi_H\} \; \text{or} \; \{\psi_H,\psi_S\}
\end{equation}
    where the robot only performs the helix motion $\psi_H$ or performs the spinning motion $\psi_S$ after $\psi_H$. Note that directly executing $\psi_S$ after grasping may not be effective since the extracting displacement of the target object is small. We propose six separation actions $a_{h},a_{hs},a_{f},a_{fs},a_{tf},a_{tfs}$ using two motion primitives and a direct lifting action $a_{dl}$. Table \ref{tab:subaction} shows their notations and illustrations. We use $M$ to represent the collection of these seven actions. 
    
    \textit{2) Action Parameter Determination:} To determine the parameters of each action and search for the best action, we define a numerical metric \textbf{action complexity} for exploring the trade-off between success rates and execution time. Let $\mathcal{A}(a)$ denote the action complexity of the action $a \in \{a_{dl},a_{h},a_{hs},a_{f},a_{fs},a_{tf},a_{tfs}\}$. It is defined by assuming that actions with larger $\theta_H$ or $\theta_S$ involve higher complexity. To reduce the search cost during exploration, we assume that the action complexity linearly scales with the success rate of each action. We find this linear relationship by executing 80 physical attempts for each action as Table \ref{tab:subaction} presents. Then, we use this hypothesis to determine the action parameters experimentally. Specifically, we predefine a set of possible values of $\theta_H,\theta_S$ experimentally for our policy to select. $\theta_H$ can be selected from $\{0,\pi,2\pi,4\pi\}$ and $\theta_S$ can be selected between $\{0,\pi/2\}$. Note that the other parameters of the motion primitives $H=(c_H,r_x,r_y,h_0,h)$ and $c_S$ are fixed in our policy. Finally, we assign integers 0 to 6 as the action complexity for the discrete actions from $a_{dl}$ to $a_{tfs}$. The action parameters and execution details are included in Table \ref{tab:subaction}.

\begin{table}[t]
\footnotesize
\vspace{0.1cm}
\renewcommand\arraystretch{0.85}
\centering
\caption{ Action Parameters and Execution Details }
\setlength\tabcolsep{1.7pt} 
    \begin{tabular}{@{\extracolsep}lccccccc}
    \toprule
    &
    $a_{dl}$ & $a_h$ & $a_{hs}$ &$a_f$& $a_{fs}$ & $a_{tf}$ & $a_{tfs}$ \\
    & 
    \includegraphics[width=10px]{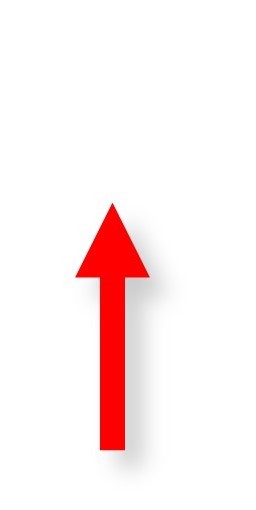} & \includegraphics[height=17px]{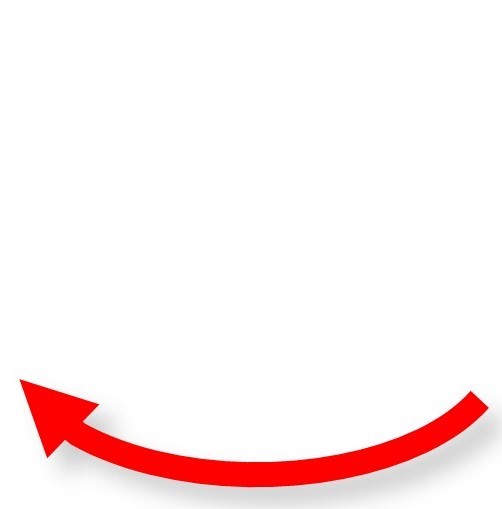} & \includegraphics[height=17px]{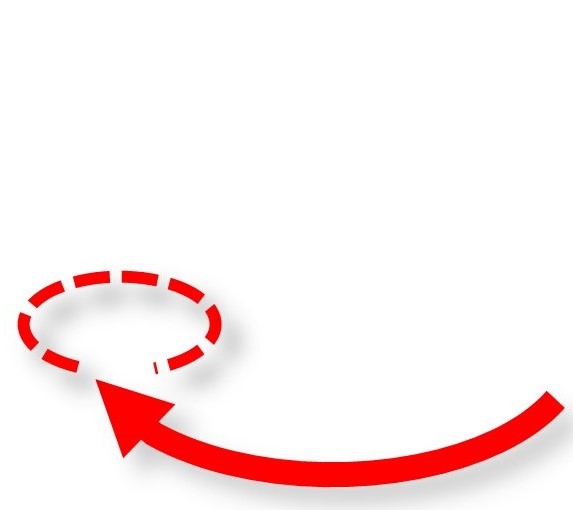} & \includegraphics[height=17px]{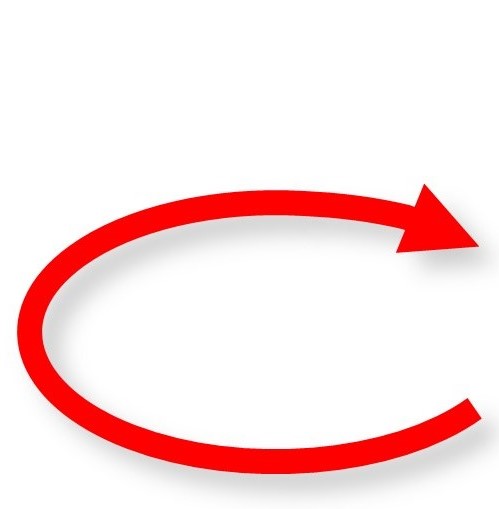} & \includegraphics[height=17px]{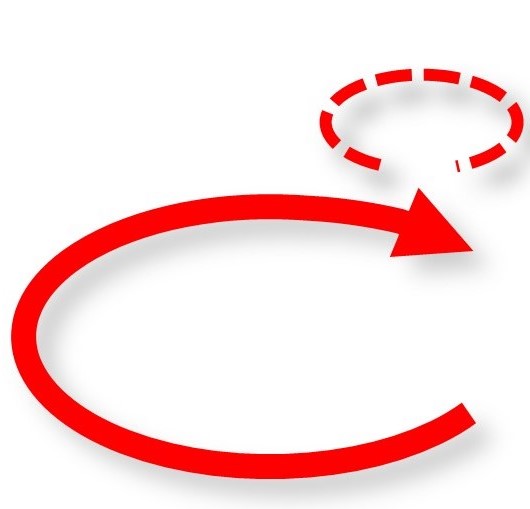} & \includegraphics[height=17px]{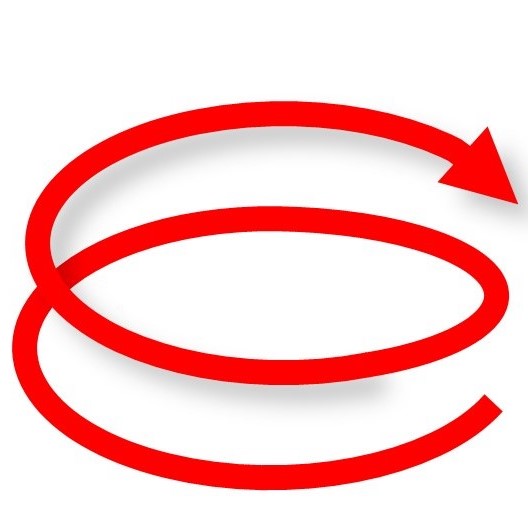} & \includegraphics[height=17px]{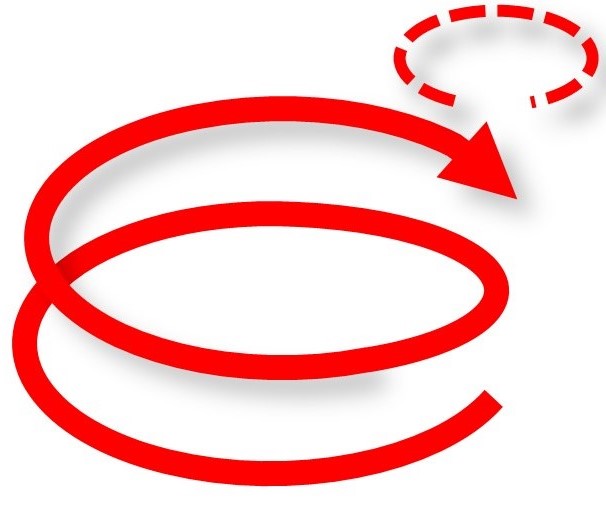} \\
    \midrule
    \textit{$\psi$} & -& $\{\psi_H\}$&$\{\psi_H,\psi_S\}$& $\{\psi_H\}$&$\{\psi_H,\psi_S\}$& $\{\psi_H\}$&$\{\psi_H,\psi_S\}$\\
    $\theta_H$& $0$ & $\pi$ & $\pi$ & $2\pi$ & $2\pi$ & $4\pi$ & $4\pi$\\
    $\theta_S$ & 0 & 0 & $\pi/2$& 0 & $\pi/2$& 0 & $\pi/2$\\
    \midrule
    Time (s)&
    1.2 & 2.3 & 2.8 & 5 & 5.5 & 8.2 & 8.7\\ 
    SR & 
    31/80 & 47/80 & 60/80 & 65/80 & 66/80 & 70/80 & 72/80\\
    \midrule
    $\mathcal{A}$ & 0 & 1 & 2 & 3 & 4 & 5 & 6\\
    \bottomrule
    \end{tabular}
\begin{tablenotes}
\item * Time (s) - Execution time of performing the action trajectory.
\item * SR - Success Rate of picking a single object.
\item * $\mathcal{A}$ - Action complexity. 
\end{tablenotes}
\label{tab:subaction}
\end{table}
    
    Our policy can explore the \textbf{optimal} action by minimizing the action complexity as much as possible. Let us consider a case when the robot performs $a_{tf}$ to extract an entangled object. Suppose the target object is entirely disentangled after a full circle ($a_{f}$) while the robot still needs to perform the second circle. Thus, the current observation only requires $a_{f}$ as the \textbf{optimal} action to ensure a successful separation with less execution time, while $a_{tf}$ is a \textbf{redundant} action which can also solve the entanglement but costs more time. We can observe that an optimal action has lower action complexity than a redundant action. Thus, the optimal action is required to untangle the target with minimal action complexity. 
    
    \textit{3) Prediction Model:} The inference of the optimal action without object models should be conditioned on the grasp locations. We propose Action Success Prediction (ASP) to predict if the action-grasp pair can successfully separate the target. ASP learns a  function parameterized by $\tau$:
\begin{equation}
    p=f_{\tau}(o,g,a)
\end{equation}
    where the input is a  depth image $o$ $\in \mathbb{R}^{224 \times 224 \times 3}$ with triplicated depth values across three channels to match with the default input size of the image encoder's backbone, a pixel-wise grasp pose $g=(u,v)$ $\in \mathbb{R}^2$, a categorical action $a$ $\in \mathbb{R}^7$ and the output is a success possibility in the range of $[0,1]$. We encode the image using a ResNet-50 backbone \cite{he2016deep}, the grasp point using a single fully-connected layer with 256 units, and the categorical action using a fully-connected layer with 14 units. Then we concatenate the output from all three branches and feed it to a fully-connected layer with 256 units and produce an action success possibility.

\begin{figure}[t]
    \centering
    \vspace{0.1cm}
    {\includegraphics[width=\linewidth]{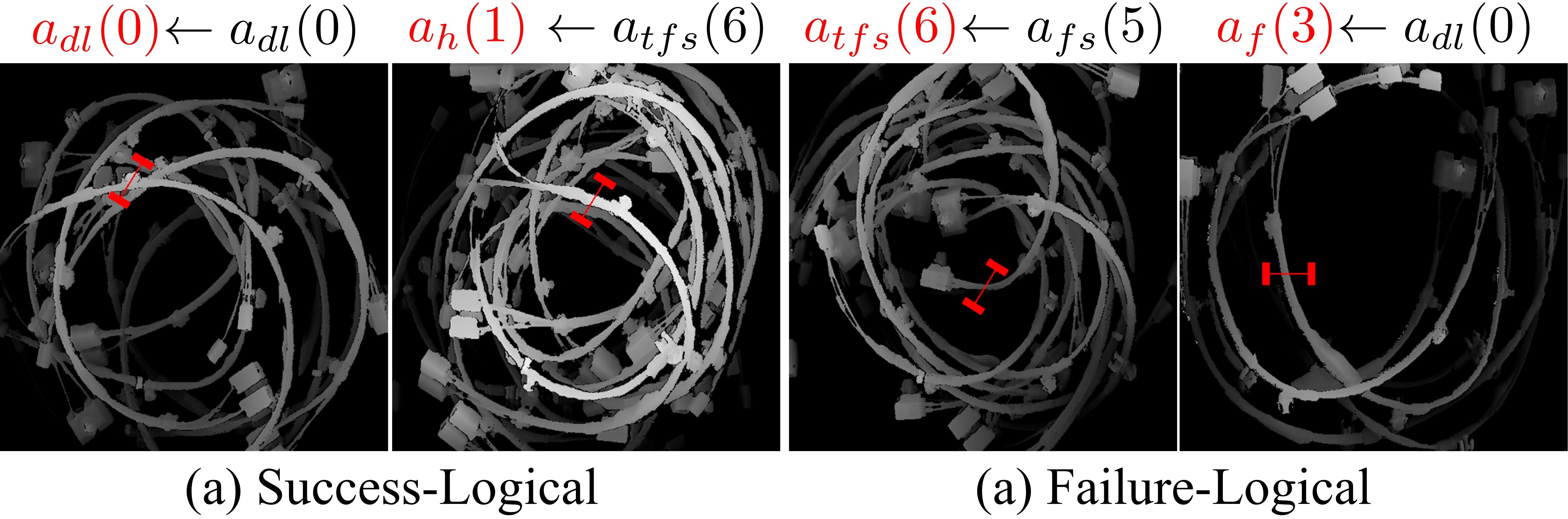}}
    \caption{Success-logical and failure-logical samples in the dataset. The notation above each figure denotes predicted action $\leftarrow$ labeled action. The numbers inside the brackets refer to the action complexity. }
    \label{fig:action-logi} 
\end{figure}   

\normalem
\setcounter{algocf}{0}
\setlength{\intextsep}{20pt} 
\begin{algorithm}[t]
\DontPrintSemicolon
    \SetKwInput{KwInput}{input} 
    \SetKwInput{KwOutput}{output} 
    \SetKwRepeat{Do}{do}{while} 
    
    \KwInput{Data pool, transfer ratio $r$, actions $M$}
    \KwOutput{ASP model $\tau$}
    
    Select training data from data pool\;
    Train ASP model $\tau$ using training data\;
    
    \While{\rm data pool is not empty}{
        $N \leftarrow$ number of samples in data pool\;
        $i \leftarrow 0$\; 
        \While{$i \leqslant r \times N$}{
            Randomly select $\{o,g,a,S\}$ from data pool\;
            $a_p \leftarrow $\texttt{ActionGraspInference}($o,g,M,\tau$)\;
            \uIf{$S=1$ \rm and $\mathcal{A}(a_p) \leqslant \mathcal{A}(a)$}{
                {\tcp{Success-logical}
                Move to training data, $i=i+1$\;}
            }
            \uElseIf{$S=0$ \rm and $\mathcal{A}(a_p) > \mathcal{A}(a)$}{
                {\tcp{Failure-logical}
                Move to training data, $i=i+1$\;}
            }
        }
        Fine-tune ASP model $\tau$ using training data\;
    }
\caption{Active Learning Algorithm}
\label{alg:asp}
\end{algorithm}

    \textit{4) Training via Active Learning:} The dataset for training ASP is entirely collected from real-world experiments. Each sample has a depth image $o$, a grasp location $g$, a labeled action $a$ and a binary success metric $S=\{0,1\}$. We execute each action for the clusters with 6, 10, 12 and 18 objects. We label each attempt with success ($S=1$) or failure ($S=0$) depending on if the robot picks a single wire harness. Due to this data collection manner, some samples in the dataset are labeled with redundant actions instead of optimal actions. To deal with this problem, we leverage active learning to train the ASP model, making it possible to predict optimal actions using this dataset. Generally, we first select several samples manually as training data to train the model, use the trained model to predict the remaining samples, query and transfer samples for training and fine-tune the model repeatedly. Specifically, we manually select the initial training data with approximately optimal actions. Note the number of samples for each action is roughly equal. Let data pool denote the left samples except for the training data. After training, we query the samples in the data pool and transfer the \textbf{logical} samples to the training data. Here, a sample $(o,g,a,S)$ can be determined as \textbf{success-logical} or \textbf{failure-logical} using the trained model $\tau$ and our proposed Action-Grasp Inference module (Section IV.C). Let $a_p=$ \texttt{ActionGraspInference}($o,g,M,\tau$) denote the predicted action: 
    \begin{itemize}
        \item Success-logical: $\mathcal{A}(a_p) \leqslant \mathcal{A}(a)$. For samples labeled with $S=1$, the labeled action $a$ is a redundant action compared with the predicted action $a_p$ (Fig. \ref{fig:action-logi}(a)).
        \item Failure-logical: $\mathcal{A}(a_p) > \mathcal{A}(a)$. For samples labeled with action $a$ and failure $S=0$, the predicted action $a_p$ has higher action complexity (Fig. \ref{fig:action-logi}(b)). 
    \end{itemize}

    During each iteration, as the number of logical samples increases, the model performance of predicting the optimal actions also improves. We define a transfer ratio $r$ representing the ratio of the number of samples that would be transferred in each iteration to the number of samples in the current data pool. The iteration stops when the data pool is empty or early stops before overfitting. Algorithm \ref{alg:asp} shows the detail of training ASP via active learning. 
    
\subsection{Action-Grasp Inference} \label{subsec:method-infer}
    At this point, we've obtained a set of grasp candidates, action candidates and the scores of each action-grasp pair. Our policy then needs to determine which action-grasp pair can be executed. This module infers all possible action-grasp pairs to guarantee a successful picking with minimal action complexity:
    \begin{equation}
        a^*, g^*=\texttt{ActionGraspInference}(o, G, M, \tau)
    \end{equation}
    where the inputs are a depth image $o$, a collection of actions $M$, grasp candidates $G$ with FGE scores from the Model-Free Grasp Detection module and ASP model $\tau$. This module first predicts the action success possibilities of all action-grasp pairs $P=f_{\tau}(o,G,M)$. If all possibilities in $P$ are lower than the threshold $p_{thld}$, which means all action-grasp pairs cannot solve the entanglement, we select the grasp with the highest FGE score and the most complex action $a_{tfs}$. Otherwise, the best solution is determined by the action-grasp pair with the lowest action complexity. If multiple grasps share the same action complexity, we select the pair with the highest FGE score. 

\section{Experiments and Results}\label{sec:exp}
    
    We conduct several real-world experiments to answer the following three questions: (1)  How does the learned ASP model perform using active learning? (Section \ref{subsec:exp-model}) (2) Does our bin picking policy perform more accurately and effectively than baselines? (Section \ref{subsec:exp-bin}) (3) How does our method qualitatively improve the performance of picking wire harnesses? (Section \ref{subsec:exp-qlt})

\subsection{ASP Model Performance}\label{subsec:exp-model}
\begin{figure}[t]
\vspace{0.1cm}
    \centering
    \includegraphics[width=\linewidth]{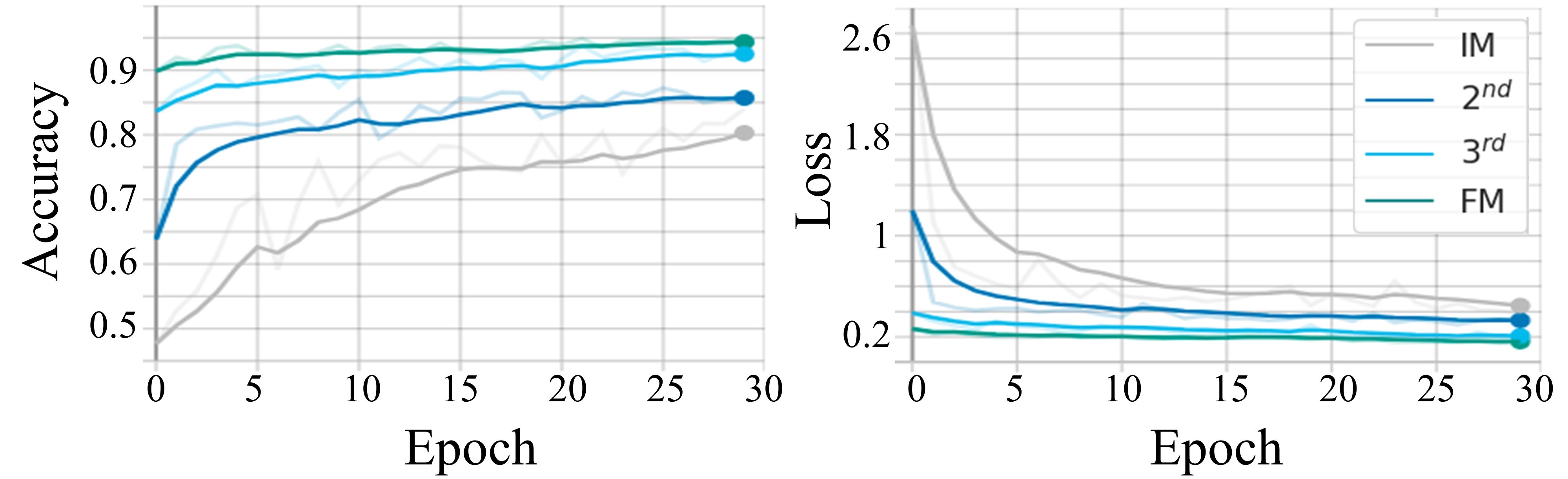} 
    \caption{Accuracy and loss of each model during action learning. }
    \label{fig:res-learn-curve} 
\end{figure}

    Our dataset contains 722 samples. We set the ratio of active learning $r=0.4$ and use a simple decision threshold of $p_{thld}=0.5$ over the softmax of each action's success possibility to classify success (1) or failure (0). We train the network using binary cross-entropy loss function and the Adam optimizer. We stop training after three times of fine-tuning as it achieves the best performance. Fig. \ref{fig:res-learn-curve} shows the accuracy and loss during active learning. The gray curve refers to the Initial Model (IM) trained using manually determined samples, which would be potentially accurate but lack robustness due to fewer data. The green line indicates the Final Model (FM), which performs the best as the fine-tuning goes on since it converges to IM but with higher data-driven accuracy. 
    
    Moreover, Table \ref{tab:active-learning} shows the details of each iteration in active learning. Row 1-2 shows the number of samples used as the training data and left in the data pool. Particularly, 92 samples left in the data pool after the final fine-tuning are used to validate all models by checking the number of logical samples. Row 3-4 shows the ratios of logical samples increase with the fine-tuning process. Finally, Table III validates our hypothesis that more complex actions correspond to higher success rates. We respectively present the average scores predicted by FM for each action. FM can correctly predict an ascending order of possibilities as the action complexity increases. We can observe that $a_{fs},a_{tf},a_{tfs}$ share similar scores since the validation samples contain 18 objects at most. $a_{tfs}$ does not show a significantly high score due to the accumulated low scores when all predictions fail and $a_{tfs}$ is forced to be selected.

\begin{table}[t] 
\footnotesize
    \centering
    \renewcommand\arraystretch{0.85}
    \begin{minipage}[c]{\linewidth}
    \vspace{-0.2cm}
        \centering
        \caption{Details and Validation Results of Active Learning}
        \begin{tabular}{lcccc}
        \toprule
        & IM & 2$\rm ^{nd}$ & 3$\rm ^{rd}$ & \textbf{FM}\\
        \midrule
        \# Samples in Training Data & 282 & 453 & 558 & 618\\
        \# Samples in Data pool & 428 & 257 & 152 & 92\\
        \midrule
        Ratio of Success-Logical ($\%$) & 78.5 & 85.7 & 87.8 & 88.9 \\
        Ratio of Failure-Logical ($\%$) & 85.1 & 80.1 & 90.1 & 91.7 \\
        \bottomrule
        \end{tabular}
        \vspace{3mm}
        \label{tab:active-learning}
    \end{minipage}
    
    \begin{minipage}[c]{\linewidth}
        \centering
        \caption{Average Predicted Scores using Validation Samples}
        \setlength\tabcolsep{5pt} 
        \begin{tabular}{cccccccc}
        \toprule
         & $a_{dl}$ & $a_h$ & $a_{hs}$ &$a_f$& $a_{fs}$ & $a_{tf}$ & $a_{tfs}$ \\
        \midrule
        $S=1$ & 0.352 & 0.489 & 0.702 & 0.750 & 0.783 & 0.730 & 0.787 \\
        $S=0$ & 0.257 & 0.375 & 0.581 & 0.636 & 0.678 & 0.606 & 0.685 \\
        \bottomrule
        \end{tabular}
        \label{tab:avg-p-val}
    \end{minipage}
\end{table}

\begin{table*}[t]
\vspace{0.1cm}
\footnotesize
    \centering
    \caption{Performance of Bin Picking Experiments}
    \renewcommand\arraystretch{0.9}
    \setlength\tabcolsep{4.8pt} 
    \begin{tabular}{llccccccccc}
    \toprule
    \multirow{2}{*}{} &
    \multirow{2}{*}{} &
    \multicolumn{3}{c}{5 Objects} &
    \multicolumn{3}{c}{10 Objects} &
    \multicolumn{3}{c}{15 Objects}\\
    \cmidrule(lr){3-5} \cmidrule(lr){6-8} \cmidrule(lr){9-11}
    & Method &Success Rate (\%) & PPH & Avg. $\mathcal{A}$ &Success Rate (\%)& PPH & Avg. $\mathcal{A}$& Success Rate (\%) & PPH& Avg. $\mathcal{A}$ \\
    \cmidrule{1-11}
    \multirow{6}{*}{Consecutive Picking}
    & DL      &  64.0 & 128 & -   &  60.0 & 92  & -   &  56.0 & 108 & -  \\
    & RAND    &  88.0 & 115 & 2.3 &  92.0 & 117 & 2.5 &  76.0 & 99  & 2.8\\
    & TFS     &  96.0 & 133 & -   &  92.0 & 127 & -   &  90.0 & 124 & -  \\
    & Ours-IM &  84.0 & 131 & 0.8 &  76.0 & 117 & 2.3 &  74.0 & 111 & 2.9\\
    & \textbf{Ours-FM} &  \text{88.0} & \textbf{156} & \textbf{0.8} &  \text{88.0} & \textbf{140} & \textbf{2.8} &  \text{86.0} & \textbf{143} & \textbf{2.3}\\
    & \textbf{Ours-FM-R} & \textbf{89.8} & \textbf{154} & \textbf{0.8} & \textbf{89.8} & \textbf{142} & \textbf{2.8} & \textbf{84.4} & \textbf{123} & \textbf{2.9}\\
    
    \midrule \midrule
    \multirow{2}{*}{} &
    \multirow{2}{*}{} &
    \multicolumn{3}{c}{18-20 Objects} &
    \multicolumn{3}{c}{20-22 Objects} &
    \multicolumn{3}{c}{22-25 Objects}\\
    \cmidrule(lr){3-5} \cmidrule(lr){6-8} \cmidrule(lr){9-11}
    & Method &Success Rate (\%) & PPH & Avg. $\mathcal{A}$ &Success Rate (\%) & PPH & Avg. $\mathcal{A}$& Success Rate (\%) & PPH& Avg. $\mathcal{A}$ \\
    \cmidrule{1-11}
    \multirow{3}{*}{Randomized Picking}
    & DL &  46.6 & 93 & - &  40.0 & 80 & - &  23.3 & 47 & -\\
    & \textbf{Ours-FM} &  \textbf{86.7}  & \textbf{113} & \textbf{2.9} &  \textbf{80.0} & \textbf{112} & \textbf{3.3} &  \textbf{73.3} & \textbf{103} & \textbf{4.3}\\
    & \textbf{Ours-FM-R} & \textbf{92.6} & \textbf{108} & \textbf{2.6} & \textbf{91.7} & \textbf{103} & \textbf{4.5} & \textbf{76.9} & \textbf{107} & \textbf{4.6}\\
    
    \bottomrule
    \end{tabular}
    \label{tab:result}
\end{table*}

\begin{figure}[t] 
    \centering
    \includegraphics[width=\linewidth]{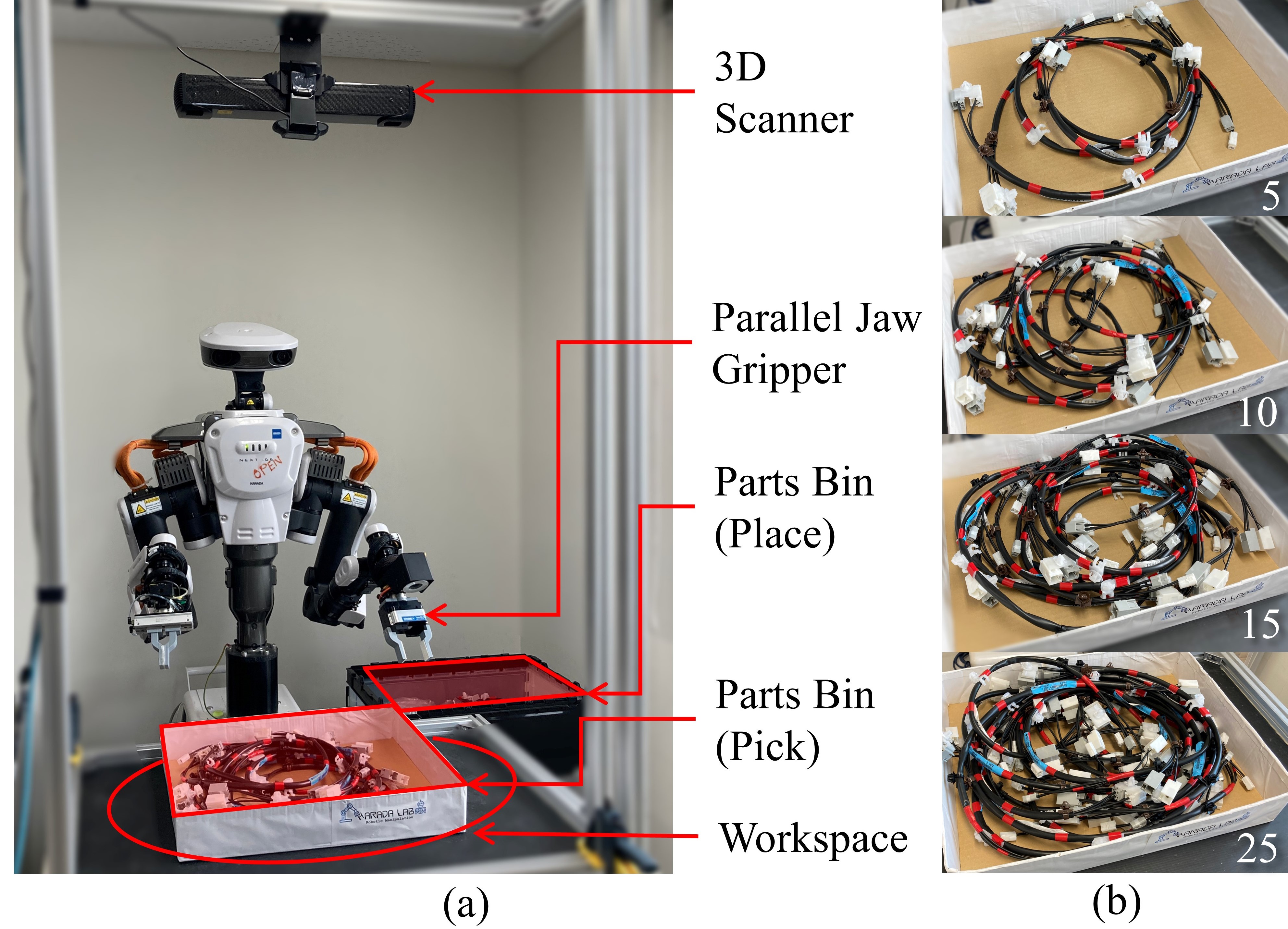} 
    \caption{Physical experiment setup for bin picking. }
    \label{fig:exp-setup} 
\end{figure}

\subsection{Bin Picking Performance}\label{subsec:exp-bin}
    
    \textit{1) Physical Experiment Setup:} We use a NEXTAGE robot from Kawada Industries Inc. The robot is required to grasp objects from the parts bin lying in front of it and transport them to another bin located on its left side. The robot's left arm operates over a workspace captured as a top-down depth image by a Photoneo PhoXi 3D scanner M. A two-fingered parallel gripper is attached at the arm tip. The setup is shown in Fig. \ref{fig:exp-setup}(a). The length of the wire harness used in this work is 74cm. After performing the analysis and physical experiments, we fix the parameters of the proposed trajectory as  $c_H=$ (0.525,0.065)[m], $r_x=$ 0.1m, $r_y=$ 0.225m, $h_0=$ 0.32m, $h=$ 0.14m as well as the speed of the action since they yield high task effectiveness. We sample several waypoints on the trajectory and plan motions with a uniform velocity. We use a PC with an Intel Core i7-CPU and 16GB memory without GPU for real-world experiments and a PC with an Intel Core i5-6400 CPU, 16GB memory and an Nvidia GeForce 1080 GPU for learning.

    We present three baselines. \textbf{DL} (directly lifting) uses FGE to detect the grasp point and executes by directly lifting ($a_{dl}$). \textbf{RAND} executes a random action and the grasp with the highest FGE score. \textbf{TFS} only executes the most complex action $a_{tfs}$ and the grasp of the highest FGE score. We also present three versions of our policy. \textbf{Ours-IM} is our policy using the initial model in active learning while \textbf{Ours-FM} uses the final model. \textbf{Ours-FM-R} denotes Ours-FM with a recovery module using force feedback. After performing the predicted action, we record the force from an F/T sensor mounted on the robot's wrist to determine if the grasped wire harness is still entangled. If there exists a sudden increase of force, the target is not disentangled and the robot places it back to the parts bin. 
    
    We leverage two metrics to evaluate the bin picking performance. \textbf{Success rate} refers to the number of successful attempts of picking up a single object divided by the number of attempts of placing. \textbf{PPH (Pickings Per Hour)} is the number of successful attempts the robot can execute in one hour. Additionally, we present \textbf{Avg. $\mathcal{A}$ (Average action complexity)} to evaluate how the action complexity predicted by our policy varies under different entanglement scenarios. 
       
    \textit{2) Task Design:} We prepare two real-world bin picking tasks. \textbf{Consecutive picking} aims to empty the bin filled with respectively 5, 10, or 15 objects. The robot picks up objects one by one until the bin is empty. \textbf{Randomized picking} refers to picking up objects from the bin filled with respectively 18-20, 20-22 and 22-25 objects. After each picking, we reload the bin and shuffle the wire harnesses to provide randomness during the task. It can encourage the robot to confront different patterns of entanglement as much as possible. Fig. \ref{fig:exp-setup}(b) shows the bins filled with different numbers of wire harnesses. 
    
\begin{figure*}[t] 
    \centering
    \vspace{0.1cm}
    \includegraphics[width=\linewidth]{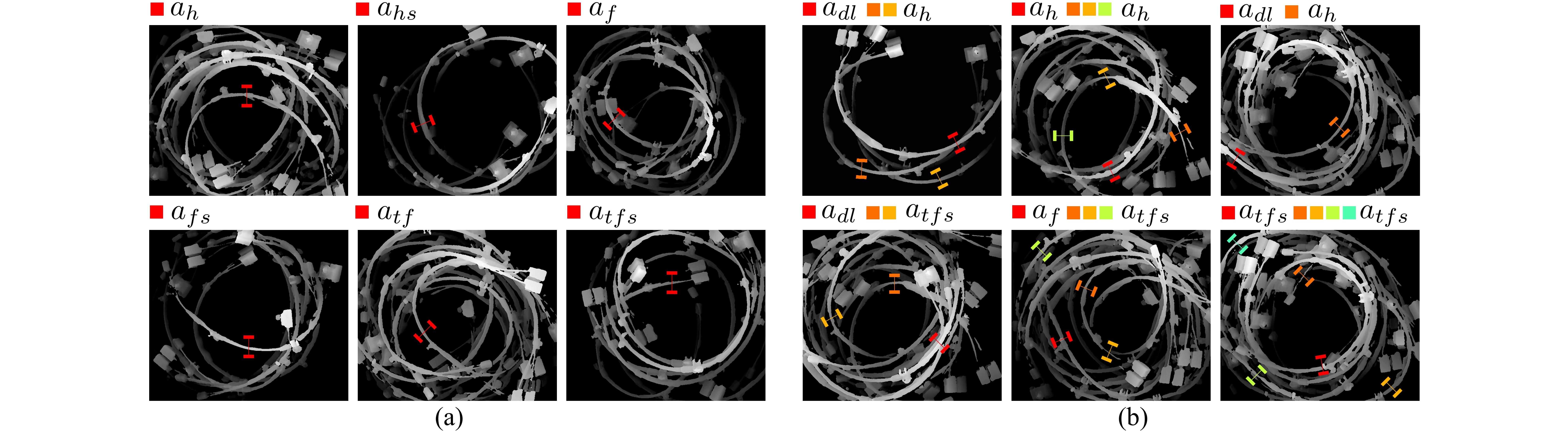} 
    \caption{Qualitative results. (a) Ours-FM predicts the optimal action-grasp pairs for each action. (b) Ours-FM predicts the best action and grasp marked using red in real-world experiments. All action-grasp pairs are presented using the same colors.}
    \label{fig:res-vis} 
\end{figure*}

    \textit{3) Comparisons with Baselines:} Table \ref{tab:result} compares the performance of the three versions of our policy and three baselines in success rate and PPH. For consecutive picking where the goal is to empty the bin,  Ours-FM and Ours-FM-R significantly increase the average success rate from 56.7\% to 87.3\% and 88.1\% compared to DL. TFS achieves higher success rates than Ours-FM but has lower PPH  since TFS only executes the time-consuming action $a_{tfs}$. Especially in the latter half of a continuous picking task when fewer objects remain in the bin, our policy can shorten the execution time by inferring adequate actions. Ours-FM-R also has lower PPH since this policy needs extra actions to place the entangled objects back in the parts bin. Furthermore, the average action complexities for the predicted actions using RAND, Ours-IM, Ours-FM and Ours-FM-R are also presented in Table \ref{tab:result}. The average action complexity for 5 objects is significantly lower than that for 10 and 15 objects. As the number of objects in the bin increases, the action complexity of the predicted action increases. It demonstrates that entanglement frequently occurs when the bin contains more objects and requires more complex actions. We also observe that the failed attempts by baselines always drag objects outside the workspace, requiring human workers to rearrange after each attempt. Our policy helps maintain a relatively clean workspace during the consecutive picking thanks to the horizontal separation and our action-grasp inference algorithm. 

    For randomized picking, we compare the performance of Ours-FM and Ours-FM-R with a DL baseline as Table \ref{tab:result} shows. More objects are involved in this task than consecutive picking. Thus, the possibilities of encountering complex entanglement patterns become higher. Ours-FM completes the task with 80\% accuracy and 109 PPH, almost twice higher than DL. The results suggest that our policy can grasp the tightly intertwined objects in dense clutter. All three proposed modules collaboratively contribute to efficient bin picking from perception to manipulation planning. However, as the number of objects increases, both metrics of Ours-FM decrease. Due to heavier occlusions and visual noise, the detected grasp candidates become fewer and some entanglement patterns can hardly be recognized from the depth image. Despite this, the most complex action $a_{tfs}$ can still strive for success. Additionally, Ours-FM-R outperforms Our-FM in success rate especially when the number of objects increases thanks to the recovery module but has lower PPH. When the bin contains more than 22 objects, Ours-FM-R shows a higher success rate and PPH than Ours-FM, indicating the feedback module can help further improve the bin picking performance. 

\subsection{Qualitative Analysis}\label{subsec:exp-qlt}
    
    \textit{1) Visualized Results:} We present visualized results of picking attempts with grasps, actions and input depth images. First, Fig. \ref{fig:res-vis}(a) presents the predicted action-grasp pairs of each action. It demonstrates that our policy infers the actions not only by analyzing the object number in the scene but also by reasoning about the occlusions around the input grasp point. Additionally, if the robot grasps close to the wire harness's end, our policy tends to predict more complex actions since this case may require the gripper to handle the length by moving a larger distance. Then, Fig. \ref{fig:res-vis}(b) shows a set of successful pickings with the reasoned action-grasp candidates ranked by descending prediction scores. The optimal action-grasp pairs inferred by our policy are marked as red. Our policy can recognize the objects barely entangled with others that only require $a_{dl}$. As for the scenes that do not contain such objects, our policy can reason the entanglement situation and predict the proper actions. When the predicted scores of all action-grasp pairs are lower than $p_{thld}$, our policy executes $a_{tfs}$ and grasp with the highest FGE score, where the target is likely on the top of the pile. 

\begin{figure}[t] 
    \centering
    \includegraphics[width=\linewidth]{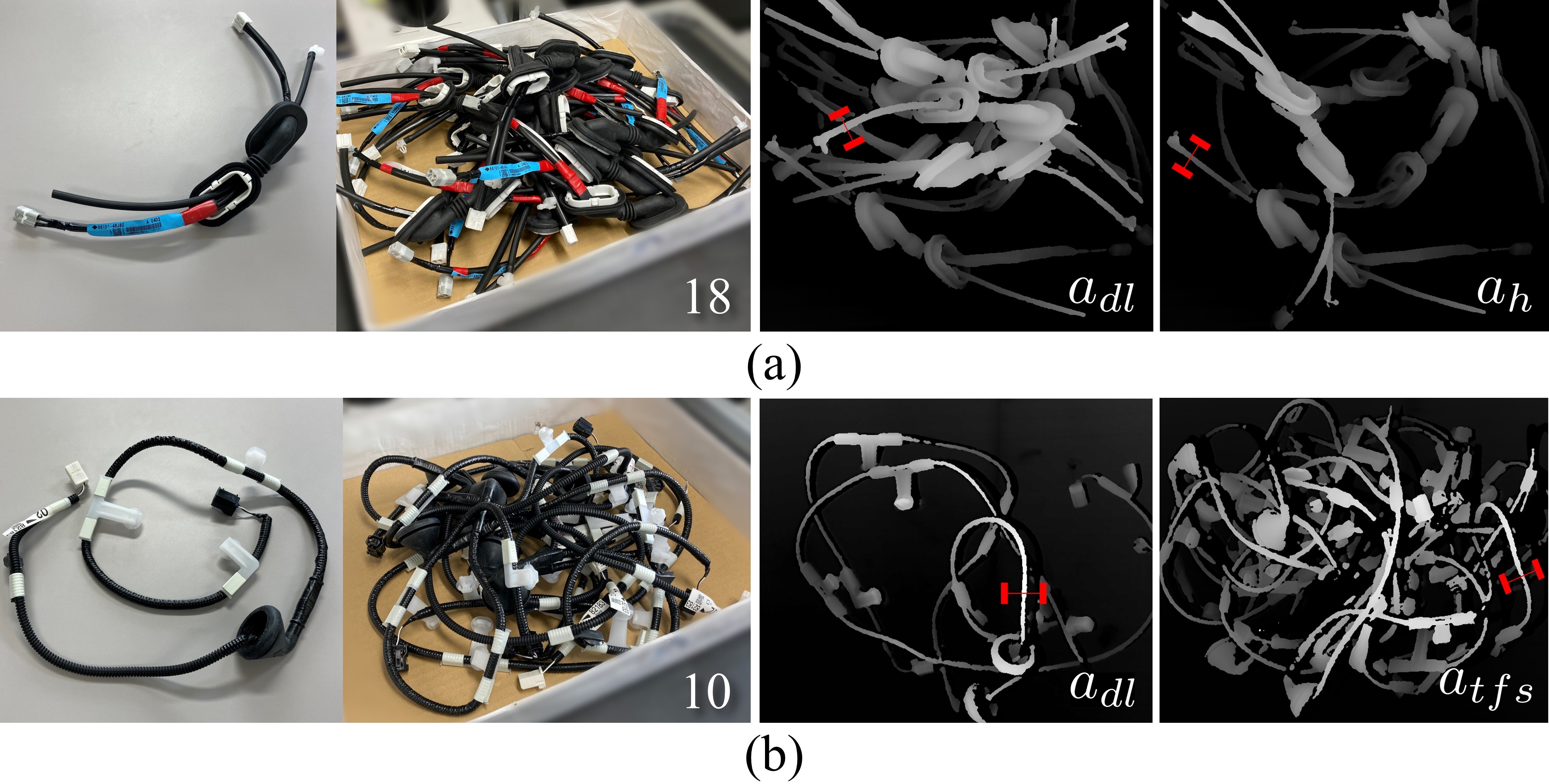}
    \caption{Novel types of wire harnesses and the predicted action-grasp pairs by our policy. (a) Short wire harnesses. (b) Long wire harnesses. }
    \label{fig:res-newobj} 
\end{figure}

    \textit{2) Novel Wire Harnesses:} To demonstrate the breadth of our method, we utilize Ours-FM for two unseen wire harnesses. They differ from those used for training in lengths and structures but have similar components (e.g., deformable cables and rigid connectors). Fig. \ref{fig:res-newobj} shows two novel wire harnesses and the corresponding action-grasp pairs predicted by our policy. Table \ref{tab:newobj} shows their length and the average action complexity of prediction with different object numbers. In the case of shorter objects (see Fig. \ref{fig:res-newobj}(a)), our model does not predict actions with too higher complexity. The robot tends to select $a_{dl}$ and $a_{h}$ to pick up objects. Since this type of wire harness is less tangle-prone, the accuracy of picking them primarily relies on the grasp detection module while our policy can handle the potential entanglement. On the other hand, for long wire harnesses (Fig. \ref{fig:res-newobj}(b)) whose length exceeds our bin picking working cell, Table \ref{tab:newobj} suggests that our policy tends to output more complex actions. However, even $a_{tfs}$ is still insufficient to separate each. More complex manipulation strategies are needed for such objects. 

\subsection{Failure Modes and Limitations}
    
    We observe four failure modes in the physical experiments.

\begin{table}[t]
\renewcommand\arraystretch{0.85}
\centering
\footnotesize
\caption{Predicted Average Action Complexity ({\rm Avg. $\mathcal{A}$)}\\ for Two Types of Unseen Wire Harnesses}
    \begin{tabular}{cccccc}
    \toprule
    Type & Length (cm) & 5 Objects & 10 Objects & 15 Objects\\
    \midrule
    Short & 45 & 0.7 & 1.3 & 1.7 \\
     Long & 115 & 4.8 & 4.6 & - \\
    \bottomrule
    \end{tabular}
\label{tab:newobj}
\end{table}

    \begin{itemize}
        \item Objects outside of the bin: The input image of the ASP model does not include the complete objects.
        \item Grasp failure: The grasp failure rate is 2.1\% (24/1170). A grasp fails when the robot grasps multiple objects in hand or grasps nothing. It mainly comes from vision sensor's noise and heavy occlusion. 
        \item Tightly wedged objects: The target tightly inserts another one's cable bundles or rigid components, making it extremely difficult to be disentangled.
        \item Action prediction failure: Our policy sometimes predicts the wrong actions for separation due to visual noise or heavily occluded objects. 
    \end{itemize}

    Our policy also has limitations. First, for long wire harnesses, the robot fails to extract them from the entanglement since their length exceeds the robot's reachable areas. Second, the training phase is unique and conditioned on the structure of the objects in the dataset. It would be difficult to adopt our current policy to wire harnesses with completely different geometries.
    
    We divide the reasons causing failure modes and limitations into two categories and provide future extensions. (1) Poor visual prediction for heavily occluded clutter: We will extend our policy by using multi-sensory inputs other than vision-only predetermined policy and force-only feedback control. We will also consider online closed-loop learning and more effective recovery methods to further improve the robustness of our policy. (2) Insufficient motion primitives: the proposed motion primitives cannot solve some complex cases and the reach range of a single robot manipulator is limited. We will consider more effective motion primitives using dual-arm or involving dynamics. It would also be interesting to design more general motion primitives to utilize our policy on various wire harnesses with different geometries.
    

\section{Conclusions}\label{sec:con}

    We present a novel bin picking system for grasping and separating entangled wire harnesses. We design an efficient post-grasping action for disentangling the target in clutter, learn a policy from real-world data to reason the extracting distance and produce the optimal action and grasp from a single depth image. Real-world experiments suggest that our policy can successfully untangle the intertwined wire harnesses from different cluttered scenes and pick them up one at a time with high accuracy. 



\normalem
\bibliographystyle{IEEEtran}
\bibliography{ebibsample.bib}


 





\end{document}